\documentclass[11pt]{article}

\usepackage[preprint]{acl}

\usepackage{microtype}
\usepackage{amsmath}
\usepackage{amssymb}
\DeclareMathOperator*{\argmax}{arg\,max}
\usepackage{multirow}
\usepackage{pdflscape}
\usepackage{graphicx}
\graphicspath{{../}}
\usepackage{tikz}
\usetikzlibrary{arrows.meta,positioning,calc,fit,backgrounds,shapes.geometric}

\usepackage[english,bidi=default]{babel} 
\babelfont{rm}{TeXGyreTermesX} 
\babelprovide[import]{hindi}
\babelfont[*devanagari]{rm}{Lohit Devanagari}
\babelprovide[import]{arabic}
\babelfont[*arabic]{rm}{Noto Sans Arabic}
\babelprovide[import, onchar=ids fonts]{chinese}  
\IfFontExistsTF{Noto Sans CJK SC}{\babelfont[chinese]{rm}[AutoFakeSlant=0.15,SmallCapsFont={Noto Sans CJK SC}]{Noto Sans CJK SC}}{}

\title{Cross-Lingual Consensus: Aligning Multilingual Cultural Knowledge via Multilingual Self-Consistency}

\author{Andrew Ivan Soegeng$^{1,2}$, \ \ Patrick Sutanto$^{2}$, \ \ Tan Sang Nguyen$^{2}$ \\
$^{1}$SAP, \quad $^{2}$School of Computing, National University of Singapore \\
\texttt{andrew.soegeng@sap.com} \\
\texttt{\{sutanto.patrick, tansang.nguyen\}@u.nus.edu}
}

\begin{document}
\maketitle
\begin{abstract}
Although Large Language Models (LLMs) demonstrate strong capabilities across various tasks, they exhibit significant performance discrepancies across languages. While prompting LLMs in English typically yields the highest general performance, it often induces a Western-centric bias, hindering the model's ability to accurately reflect diverse cultural knowledge. We hypothesize that LLMs already possess rich cultural knowledge embedded within local-language representations, but fail to retrieve it when prompted in English. To bridge this cross-lingual knowledge gap, we propose a novel self-supervised framework. Our method leverages multilingual self-consistency to identify the most reliable cultural responses across languages, combined with a self-critique mechanism to transfer this knowledge to the weaker language. Evaluations on the BLEnD benchmark demonstrate that our approach significantly improves cultural alignment—boosting performance on English queries by an average of 5.03\%—relying entirely on self-generated data. Ultimately, our work demonstrates that latent cultural knowledge can be successfully surfaced and propagated across languages, enabling more culturally equitable and consistent LLMs.

\end{abstract}

\section{Introduction}
Large Language Models (LLMs) have achieved remarkable progress across diverse natural language processing tasks, including logical reasoning \cite{mondorf2024accuracyevaluatingreasoningbehavior}, question answering \cite{NEURIPS2020_1457c0d6}, and multilingual understanding \cite{workshop2023bloom176bparameteropenaccessmultilingual}. Nevertheless, despite their strong overall performance, these models often exhibit uneven behavior across different languages \cite{DBLP:conf/emnlp/HuangTZZSXW23} and struggle to generate culturally appropriate responses when applied beyond dominant linguistic and cultural contexts \cite{naous-etal-2024-beer}.

A key challenge for multilingual LLMs is their inability to access knowledge across languages consistently. Although models often achieve the strongest performance when prompted in English, this can introduce Western-centric bias and limit their ability to reflect diverse cultural knowledge. Interestingly, the same models may produce more culturally appropriate responses when prompted in local languages \cite{ying-etal-2025-disentangling, myung2024blend}, suggesting that relevant knowledge is already present but not effectively retrieved or transferred across languages. This mismatch limits the reliability of LLMs in diverse real-world settings.

To address these limitations, prior work has explored several directions for improving cross-lingual and cultural alignment in LLMs. Prompt-based methods attempt to inject cultural knowledge at inference time \cite{wang-etal-2024-countries}, but often require careful design and fail to capture deeper cultural understanding \cite{durmustowards, kovavc2023large}. Training-based approaches instead rely on curated datasets from surveys, social media, or multilingual sources to enhance cultural awareness \cite{li2024culturellm, shi-etal-2024-culturebank, adilazuarda-etal-2025-surveys}, though these methods are costly and difficult to scale. More recent work leverages critique-based data synthesis and self-consistency to improve cultural knowledge and model reliability \cite{feng-etal-2025-culfit, wang-etal-2025-calm}. However, these approaches either depend on stronger external models or are primarily evaluated in structured settings, leaving open challenges for scalable and robust alignment in open-ended generation.

In this work, we propose a novel framework that leverages multilingual self-consistency to improve cultural knowledge alignment across languages. Instead of relying on external annotations, our method exploits the model’s own responses across multiple languages to identify reliable knowledge. By comparing response consistency across languages, we can determine which language yields more stable and coherent answers and use this signal to construct self-supervised training data.


Our contributions can be summarized as follows:
\begin{itemize}
    \item We propose a self-supervised multilingual self-consistency framework to generate reliable training signals without human annotation.
    \item We introduce a cross-lingual knowledge transfer mechanism that leverages stronger-language responses to improve weaker-language performance.
\end{itemize}

\section{Related Works}
\textbf{Cross-Lingual Performance Disparities}
Multilingual Large Language Models (LLMs) have been shown to yield different answers when a query is posed in different languages, leading to high performance variance across languages \citep{xuan-etal-2025-mmlu, bandarkar-etal-2024-belebele, ponti-etal-2020-xcopa}. This discrepancy is primarily attributed to English-skewed training data, which causes models to route their internal reasoning through English \citep{Weihua_Huang_Liu_Vangani_Zou_Tao_Wu_Aw_Chen_Lee_2026, schut2025multilingual, wendler-etal-2024-llamas}. Consequently, as the target language drifts further from English, models suffer significant performance degradation, particularly in low-resource languages \citep{huang-etal-2024-1}. Conversely, in some cases, LLMs may possess cultural knowledge in a local language but fail to retrieve or translate it when prompted in English \citep{myung2024blend}.

\textbf{Cultural Bias and Western-Centricity}
Cultural bias has emerged as a significant concern in current LLMs as these models tend to reflect the dominant values of their pretraining corpora, often marginalizing other demographic groups \citep{liattributing}. Consequently, LLMs predominantly exhibit a bias toward Western values \citep{mushtaq2025towards, liculture}. Moreover, simply prompting these models to adopt a specific cultural perspective often yields answers grounded in superficial stereotypes rather than a deep understanding of the underlying cultural nuances \citep{durmustowards}. Such biases raise serious safety concerns, particularly regarding the deployment of LLMs in underrepresented or localized cultural contexts \citep{azmi-etal-2025-indosafety}.

\textbf{Approaches to Cultural Alignment}
Various strategies attempt to mitigate cultural bias in LLMs. One approach aims to directly inject cultural knowledge via prompting \citep{wang-etal-2024-countries}. However, such methods often yield only a shallow understanding and require extensive domain expertise to design \citep{durmustowards, kovavc2023large}. Other approaches fine-tune models to improve their awareness of specific cultures. These methods often involve curating training datasets from surveys \citep{li2024culturellm}, social media \citep{shi-etal-2024-culturebank}, or a combination of diverse sources \citep{adilazuarda-etal-2025-surveys}. Recent work also leverages stronger LLMs to generate critiques that improve cultural data quality \citep{feng-etal-2025-culfit}. Another line of research demonstrates that self-supervision via self-consistency can enhance cultural knowledge, even without relying on stronger models \citep{wang-etal-2025-calm, zhang-etal-2025-cm}. Building on these foundations, our work demonstrates how to further enhance cultural alignment in open-ended generation by combining self-critique and self-consistency across languages, eliminating the reliance on stronger LLMs.

\section{Methodology}
We adapt CulFiT \cite{feng-etal-2025-culfit} by omitting the ineffective Direct Preference Optimization (DPO) phase and introducing self-supervised ground truth generation. The pipeline has two stages: (1) \textit{Bilingual Question Generation} for synthesizing English and local-language query pairs, and (2) \textit{Self-Supervised Ground Truth Generation} to distill reliable cultural knowledge from a base model $\mathcal{M}$ via multilingual self-consistency.

\begin{figure*}[t]
\centering
\includegraphics[width=\textwidth]{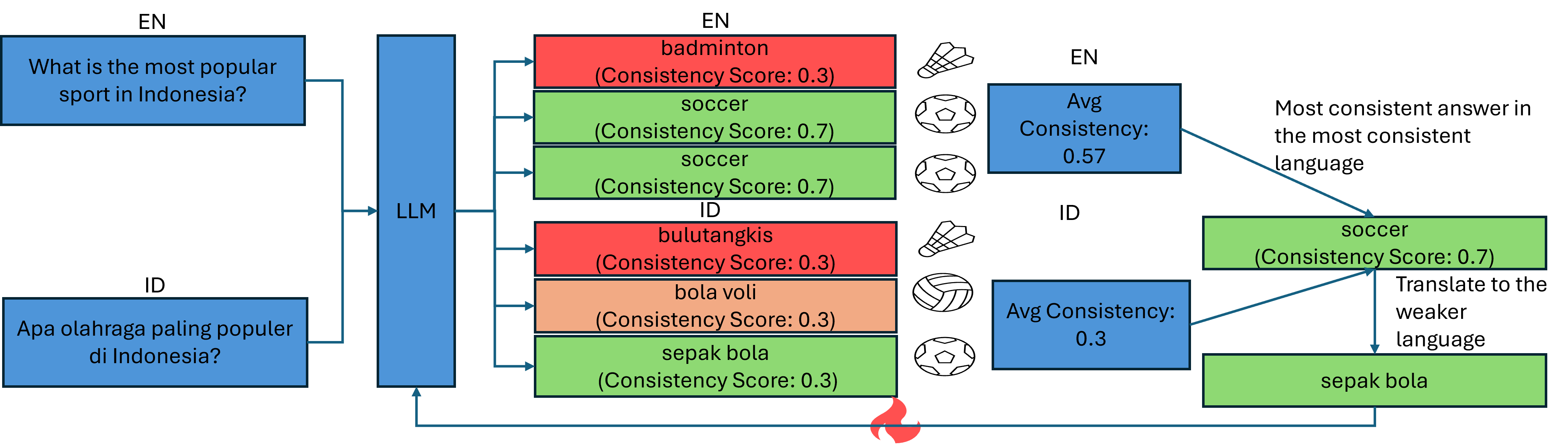}
\caption{Overview of the self-supervised ground truth generation via multilingual self-consistency. The model generates $N$ responses per language; the language with higher intra-language consistency is selected as the stronger language, and the most consistent answer in the stronger language is translated to the weaker language and set as the ground truth.}
\label{fig:method}
\end{figure*}

\subsection{Bilingual Question Generation}
We convert assertive statements $s_i$ from the CANDLE \cite{10.1145/3543507.3583535} and CultureAtlas \cite{fung2024massivelymulticulturalknowledgeacquisition} datasets into coherent knowledge paragraphs $p_i$ by prompting $\mathcal{M}$. We then prompt $\mathcal{M}$ to generate culturally grounded questions $\{q_1, \ldots, q_K\}$ from $p_i$ and extract each question's country of origin $o_k$ and primary language $\ell_k$. Retaining only non-English questions supported by Google Translate, we translate each into the local language to form bilingual pairs:
\begin{equation}
    \mathcal{Q} = \left\{ \left( q_k^{\text{en}},\; q_k^{\ell_k} \right) \mid k = 1, \ldots, |\mathcal{Q}| \right\}
\end{equation}
where $q_k^{\text{en}}$ and $q_k^{\ell_k}$ are the English and local language translations.

\subsection{Self-Supervised Ground Truth Generation}
Given $\mathcal{Q}$, we generate self-supervised ground truth answers based on $\mathcal{M}$'s response consistency.

\paragraph{Response Sampling.} For each $q_k$ and language $\lambda \in \{\text{en},\; \ell_k\}$, we sample $N$ independent responses from $\mathcal{M}$:
\begin{equation}
    \mathcal{A}_k^{\lambda} = \left\{ a_{k,1}^{\lambda},\; \ldots,\; a_{k,N}^{\lambda} \right\}
\end{equation}

\paragraph{Intra-Language Consistency.} We evaluate internal agreement within each language's responses using pairwise cosine similarity of Qwen3-Embedding-0.6B \cite{qwen3embedding} embeddings, $\mathbf{e}(\cdot)$. The consistency score $C_k^{\lambda}$ is the average similarity over all $\binom{N}{2}$ unique pairs:
\begin{equation}
    C_k^{\lambda} = \frac{2}{N(N-1)} \sum_{1 \leq i < j \leq N} \cos\!\left( \mathbf{e}(a_{k,i}^{\lambda}),\; \mathbf{e}(a_{k,j}^{\lambda}) \right)
\end{equation}

\paragraph{Ground Truth Selection.} The language with higher consistency ($C_k^{\lambda}$) is designated the \textit{stronger} language $\lambda^{+}$; the other is the \textit{weaker} language $\lambda^{-}$. The self-supervised ground truth $g_k^{*}$ is the $\lambda^{+}$ response with the highest average pairwise similarity to its peers:
\begin{equation}
    g_k^{*} = \argmax_{a_{k,i}^{\lambda^{+}}} \frac{1}{N-1} \sum_{\substack{j=1 \\ j \neq i}}^{N} \cos\!\left( \mathbf{e}(a_{k,i}^{\lambda^{+}}),\; \mathbf{e}(a_{k,j}^{\lambda^{+}}) \right)
\end{equation}
Next, we translate $g_k^{*}$ into $\lambda^{-}$ via Google Translate, producing $\hat{g}_k^{\lambda^{-}}$ and use that as the ground truth to improve the model on the weaker language.

\paragraph{Critique-Augmented Training Targets.} Following \cite{feng-etal-2025-culfit}, we sample a response $m_k$ from $\mathcal{A}_k^{\lambda^{-}}$ and prompt $\mathcal{M}$ to generate a critique $c_k$ comparing $m_k$ against $\hat{g}_k^{\lambda^{-}}$.

\paragraph{Training Data Construction.} Each instance comprises the tuple $(q_k^{\lambda^{-}}, m_k, c_k, \hat{g}_k^{\lambda^{-}})$. We restrict the dataset to instances associated with the 16 cultural regions defined in the BLEnD benchmark. To mitigate catastrophic forgetting, we augment this corpus with the Aya Dataset \cite{singh-etal-2024-aya}, maintaining a 3:1 ratio of cultural to general examples.

\section{Experimental Setup}

\begin{table*}[t]
\centering
\footnotesize
\setlength{\tabcolsep}{1.5pt}
\begin{tabular}{l|cccccccccccccccc|c}
\hline
\textbf{Model} & \textbf{UK} & \textbf{US} & \textbf{CN} & \textbf{ES} & \textbf{MX} & \textbf{DZ} & \textbf{GR} & \textbf{KR} & \textbf{JB} & \textbf{IR} & \textbf{ID} & \textbf{AZ} & \textbf{KP} & \textbf{NG} & \textbf{AS} & \textbf{ET} & \textbf{Avg} \\
\hline
\multicolumn{18}{c}{\textbf{English}} \\
\hline
Llama 3.1 & 78.95 & 82.86 & 62.66 & 52.67 & 58.46 & 49.89 & 55.87 & 52.00 & 37.55 & 57.24 & 55.42 & 52.88 & 41.89 & 35.38 & 42.56 & 40.68 & 53.56 \\
Ours & \textbf{82.24} & \textbf{85.03} & \textbf{66.46} & \textbf{61.62} & \textbf{65.76} & \textbf{56.06} & \textbf{59.13} & \textbf{59.37} & \textbf{45.20} & 60.09 & \textbf{61.67} & \textbf{59.70} & \textbf{49.55} & \textbf{38.90} & \textbf{47.38} & \textbf{44.70} & \textbf{58.93} \\
\hline
\multicolumn{18}{c}{\textbf{Local}} \\
\hline
Llama 3.1 & 78.95 & 82.86 & \textbf{60.13} & 62.90 & 63.88 & 42.11 & 38.04 & 53.68 & \textbf{28.60} & 43.86 & 53.12 & 40.72 & 37.16 & \textbf{24.62} & \textbf{16.56} & \textbf{4.24} & 45.72 \\
Ours & \textbf{82.24} & \textbf{85.03} & 55.70 & \textbf{72.28} & \textbf{69.31} & 43.25 & 39.13 & 52.21 & 25.11 & 46.05 & 51.04 & 39.66 & 36.49 & 21.32 & 13.42 & 1.69 & 45.87 \\
\hline
\end{tabular}
\caption{BLEnD evaluation results (SEM-B scores) comparing Llama~3.1~8B~Instruct (baseline) and our method across 16 countries/regions (see Table~\ref{tab:country_codes} in the Appendix for country code mappings). \textbf{Bold} values indicates a statistically significantly better model ($p < 0.05$)}
\label{tab:blend_results}
\end{table*}

\begin{table}[t]
\centering
\small
\begin{tabular}{lcc}
\hline
\textbf{Benchmark} & \textbf{Llama 3.1} & \textbf{Ours} \\
\hline
M-HellaSwag (30 languages) & \textbf{49.29} & 48.44 \\
Global MMLU (15 languages) & \textbf{54.77} & 53.35 \\
\hline
\end{tabular}
\caption{Macro-averaged accuracy (\%) on general reasoning benchmarks.}
\label{tab:general_results}
\end{table}

\subsection{Base Model and Target Languages}
We use Llama~3.1~8B~Instruct \cite{grattafiori2024llama3herdmodels} ($\mathcal{M}$) for training data synthesis and fine-tuning. We target the 13 languages (English, Chinese, Spanish, Indonesian, Korean, Greek, Persian, Arabic, Azerbaijani, Sundanese, Assamese, Hausa, and Amharic) and 16 regions of the BLEnD benchmark \cite{myung2024blend}. The pipeline yields 5{,}007 cultural instances, combined with 1{,}668 Aya Dataset instances (6{,}675 total). Appendix~\ref{tab:training_data} details the cultural data distribution. Hyperparameters are in Appendix~\ref{app:training}.

\subsection{Evaluation}
We evaluate cultural alignment using BLEnD \cite{myung2024blend} (52.6k QA pairs across 16 regions in English and local languages). We assess general commonsense and multilingual reasoning preservation using Multilingual HellaSwag \cite{lai-etal-2023-okapi} and Global MMLU \cite{singh2024globalmmlu}. Significance is reported via paired bootstrap resampling ($p < 0.05$).

\section{Results}
\subsection{Main Results} 
Our method significantly improves English-setting performance across all 16 regions (+5.03\% average) (Table~\ref{tab:blend_results}). However, severe data scarcity (<1\% of total cultural data) for low-resource languages (Hausa, Assamese, Sundanese) caused catastrophic forgetting, reducing local-language performance in these regions. 

General capabilities remain largely intact (Table~\ref{tab:general_results}), with minor decreases on Multilingual HellaSwag (-0.85\%) and Global MMLU (-1.42\%). Per-language breakdowns are in Appendices~\ref{tab:hellaswag_full} and~\ref{tab:gmmlu_full}.

\subsection{Ablation Study} 

\begin{table}[t]
\centering
\small
\begin{tabular}{l c c c c}
\hline
\textbf{Model} & \textbf{Filter} & \textbf{Consistency} & \textbf{Avg Local} & \textbf{Avg EN} \\
\hline
Llama 3.1 & -- & -- & 45.72 & 53.56 \\
\hline
\multirow{4}{*}{Ours} 
     &     --     &    --      & 45.85 & 55.44 \\
     &     --     & \checkmark & 44.72 & 58.89 \\
     & \checkmark &    --      & \textbf{45.92} & 55.86 \\
     & \checkmark & \checkmark & 45.87 & \textbf{58.93} \\
\hline
\end{tabular}
\caption{Ablation results evaluating the impact of the filter and consistency components. Active components are indicated with a checkmark (\checkmark). Scores represent averaged accuracy (\%).}
\label{tab:ablation_results}
\end{table}

We perform an ablation study to evaluate the effectiveness of our proposed ground-truth selection mechanism, which relies on cross-lingual self-consistency. We compare this against a variant where the ground-truth response is selected randomly. Additionally, we analyze the impact of filtering the training data to include only the specific languages evaluated in our study. The results are presented in Table \ref{tab:ablation_results}.

First, our findings show that filtering the data to our target languages does not degrade overall performance, while significantly improving training efficiency by reducing the dataset size. More importantly, removing the self-consistency module (i.e., using random selection) results in a substantial performance drop in the English setting compared to our full method. Despite this degradation, the random-selection model still outperforms the unaligned base model. We attribute this underlying performance gain to the critique-augmented training format, which has been demonstrated to be a crucial component for alignment in prior work \citep{feng-etal-2025-culfit}.

\section{Conclusion}
In this work, we propose a self-supervised framework for improving cross-lingual cultural alignment in large language models through multilingual self-consistency. By leveraging the model’s own responses across languages, our approach identifies reliable knowledge and transfers it from stronger to weaker languages without requiring external annotations or stronger teacher models. Experimental results on the BLEnD benchmark demonstrate that our method significantly improves cultural alignment, in English settings, while largely preserving general reasoning capabilities. 

Despite these gains, our analysis reveals limitations in low-resource languages, where data scarcity can lead to performance degradation. This suggests that while self-supervision is effective, it remains sensitive to the availability and balance of multilingual data. Overall, our findings highlight the potential of exploiting latent cross-lingual knowledge within LLMs to achieve more culturally consistent and equitable behavior.

\bibliography{custom}

\appendix

\section{Training Details}
\label{app:training}
We fine-tune $\mathcal{M}$ using LoRA \cite{hu2022lora} applied to all linear layers with rank $r = 16$. Training is conducted with a learning rate of $1 \times 10^{-5}$ using a cosine learning rate scheduler with a warmup ratio of 0.1. We use a per-device batch size of 2 with no gradient accumulation with 8 H200 GPUs, and train for 1{,}000 steps with a maximum sequence length of 4{,}096 tokens. Training is performed in bfloat16 precision.

\section{Multi-Turn Dialogue Format}
\label{app:dialogue}
Each training instance is structured as a six-turn dialogue between a user and an assistant. The cross-entropy loss during supervised fine-tuning is computed exclusively over the assistant turns (turns 2, 4, and 6), with all user turns masked.
\begin{enumerate}
    \item \textbf{User:} $q_k^{\lambda^{-}}$ \hfill {\small(culturally grounded question)}
    \item \textbf{Assistant:} $m_k$ \hfill {\small(sampled answer in weaker language)}
    \item \textbf{User:} Critique request \hfill {\small(prompt to identify errors)}
    \item \textbf{Assistant:} $c_k$ \hfill {\small(critique of $m_k$)}
    \item \textbf{User:} Refinement request \hfill {\small(prompt to correct the answer)}
    \item \textbf{Assistant:} $\hat{g}_k^{\lambda^{-}}$ \hfill {\small(translated ground truth)}
\end{enumerate}
This structure guides the model through a self-reflective reasoning process: first reproducing a potentially flawed response, then identifying its shortcomings, and finally producing a corrected answer grounded in cross-lingual consensus.

\begin{table}[t]
\centering
\small
\begin{tabular}{llrrr}
\hline
\textbf{Code} & \textbf{Region} & \textbf{En} & \textbf{Local} & \textbf{Total} \\
\hline
DZ & Algeria & 33 & 1{,}315 & 1{,}348 \\
ES, MX & Spain, Mexico & 558 & 711 & 1{,}269 \\
CN & China & 144 & 739 & 883 \\
IR & Iran & 34 & 350 & 384 \\
GR & Greece & 52 & 327 & 379 \\
KR, KP & S.~Korea, N.~Korea & 38 & 323 & 361 \\
ID & Indonesia & 115 & 90 & 205 \\
ET & Ethiopia & 4 & 87 & 91 \\
NG & N.~Nigeria & 3 & 69 & 72 \\
AZ & Azerbaijan & 0 & 13 & 13 \\
AS & Assam & 0 & 1 & 1 \\
JB & West Java & 0 & 1 & 1 \\
\hline
& \textbf{Total} & \textbf{981} & \textbf{4{,}026} & \textbf{5{,}007} \\
\hline
\end{tabular}
\caption{Distribution of synthesized cultural training instances across regions, sorted by total count.}
\label{tab:training_data}
\end{table}

\begin{table}[h]
\centering
\small
\begin{tabular}{ll|ll}
\hline
\textbf{Code} & \textbf{Country/Region} & \textbf{Code} & \textbf{Country/Region} \\
\hline
UK & United Kingdom & ID & Indonesia \\
US & United States & AZ & Azerbaijan \\
CN & China & KP & North Korea \\
ES & Spain & NG & Northern Nigeria \\
MX & Mexico & AS & Assam \\
DZ & Algeria & ET & Ethiopia \\
GR & Greece & JB & Western Java \\
KR & South Korea & IR & Iran \\
\hline
\end{tabular}
\caption{Country/region codes used in Table~\ref{tab:blend_results}.}
\label{tab:country_codes}
\end{table}

\begin{table}[h]
\centering
\small
\setlength{\tabcolsep}{3pt}
\begin{tabular}{lccr}
\hline
\textbf{Lang} & \textbf{Llama 3.1} & \textbf{Ours} & \textbf{Diff} \\
\hline
ar & \textbf{48.96} & 47.46 & $-$1.50 \\
bn & \textbf{36.68} & 36.27 & $-$0.41 \\
ca & \textbf{58.27} & 57.17 & $-$1.10 \\
da & \textbf{58.02} & 56.98 & $-$1.04 \\
de & \textbf{61.22} & 60.24 & $-$0.98 \\
es & \textbf{67.75} & 66.46 & $-$1.29 \\
eu & \textbf{32.72} & 32.30 & $-$0.42 \\
fr & \textbf{65.76} & 64.91 & $-$0.86 \\
gu & \textbf{34.76} & 34.32 & $-$0.44 \\
hi & \textbf{45.03} & 43.77 & $-$1.26 \\
hr & \textbf{50.97} & 50.06 & $-$0.91 \\
hu & \textbf{50.83} & 49.73 & $-$1.10 \\
hy & 29.48 & \textbf{29.68} & +0.20 \\
id & \textbf{59.42} & 58.43 & $-$0.99 \\
it & \textbf{64.10} & 62.86 & $-$1.24 \\
kn & 33.47 & \textbf{33.30} & $-$0.17 \\
ml & 31.29 & \textbf{31.10} & $-$0.19 \\
mr & \textbf{34.82} & 34.24 & $-$0.58 \\
ne & \textbf{35.30} & 34.44 & $-$0.86 \\
nl & \textbf{62.55} & 61.59 & $-$0.96 \\
pt & \textbf{65.61} & 64.33 & $-$1.28 \\
ro & \textbf{58.22} & 56.73 & $-$1.48 \\
ru & \textbf{58.85} & 58.02 & $-$0.83 \\
sk & \textbf{50.02} & 49.19 & $-$0.82 \\
sr & \textbf{50.95} & 50.00 & $-$0.95 \\
sv & \textbf{60.82} & 59.65 & $-$1.17 \\
ta & 30.55 & \textbf{30.68} & +0.13 \\
te & \textbf{31.91} & 31.71 & $-$0.21 \\
uk & \textbf{53.96} & 52.72 & $-$1.24 \\
vi & \textbf{56.39} & 54.97 & $-$1.42 \\
\hline
\textbf{Avg} & \textbf{49.29} & 48.44 & $-$0.85 \\
\hline
\end{tabular}
\caption{Multilingual HellaSwag per-language accuracy (\%).}
\label{tab:hellaswag_full}
\end{table}

\begin{table}[h]
\centering
\small
\setlength{\tabcolsep}{3pt}
\begin{tabular}{lccr}
\hline
\textbf{Lang} & \textbf{Llama 3.1} & \textbf{Ours} & \textbf{Diff} \\
\hline
ar & \textbf{53.25} & 47.00 & $-$6.25 \\
bn & 46.00 & \textbf{46.25} & +0.25 \\
de & \textbf{59.75} & 58.75 & $-$1.00 \\
en & \textbf{67.25} & 66.75 & $-$0.50 \\
es & \textbf{60.25} & 58.25 & $-$2.00 \\
fr & \textbf{58.50} & 58.25 & $-$0.25 \\
hi & \textbf{48.00} & 47.00 & $-$1.00 \\
id & 58.00 & 58.00 & 0.00 \\
it & \textbf{61.00} & 60.50 & $-$0.50 \\
ja & \textbf{56.75} & 54.25 & $-$2.50 \\
ko & \textbf{52.00} & 51.50 & $-$0.50 \\
pt & \textbf{62.25} & 60.00 & $-$2.25 \\
sw & \textbf{44.00} & 42.75 & $-$1.25 \\
yo & \textbf{34.75} & 33.25 & $-$1.50 \\
zh & \textbf{59.75} & 57.75 & $-$2.00 \\
\hline
\textbf{Avg} & \textbf{54.77} & 53.35 & $-$1.42 \\
\hline
\end{tabular}
\caption{Global MMLU per-language accuracy (\%).}
\label{tab:gmmlu_full}
\end{table}

\end{document}